# Untethered thin dielectric elastomer actuated soft robot

Xi Wang[1,2], Jing Liu[1,2], Siqian Li[1,2], Hengtai Dai[1,2], Jung-Che Chang[1,2], Dragos Axinte[1,2], Xin Dong[1,2]*

[1]Rolls-Royce University Technology Centre in Manufacturing and On-Wing Technology, Faculty of Engineering, University of Nottingham, UK, NG7 2GX

[2]Department of mechanical materials and manufacturing engineering, Faculty of Engineering, University of Nottingham, UK, NG8 1BB

*Corresponding Author: Xin Dong (email: xin.dong@nottingham.ac.uk).

*The research leading to these results has received funding from China Scholarship Council and the Industrial Strategy Challenge Fund delivered by UK Research and managed by EPSRC under Grant Agreement No. EP/R026084/1.

**Abstract**

Thin dielectric elastomer actuator (DEA) features a unique in-plane configuration, enabling low-profile designs capable of accessing millimetre-scale narrow spaces. However, most existing DEA-powered soft robots require high voltages and wired power connections, limiting their ability to operate in confined environments. This study presents an untethered thin soft robot (UTS-Robot) powered by thin dielectric elastomer actuators (TS-DEA). The robot measures 38 mm in length, 6 mm in height, and weighs just 2.34 grams, integrating flexible onboard electronics to achieve fully untethered actuation. The TS-DEA, operating at resonant frequencies of 86 Hz under a low driving voltage of 220 V, adopts a dual-actuation sandwiched structure, comprising four dielectric elastomer layers bonded to a compressible tensioning mechanism at its core. This design enables high power density actuation and locomotion via three directional friction pads. The low-voltage actuation is achieved by fabricating each elastomer layer via spin coating to an initial thickness of 50 μm, followed by biaxial stretching to 8 μm. A comprehensive design and modelling framework has been developed to optimise TS-DEA performance. Experimental evaluations demonstrate that the bare TS-DEA achieves a locomotion speed of 12.36 mm/s at resonance, the untethered configuration achieves a locomotion speed of 0.5 mm/s, making it highly suitable for navigating confined and complex environments.

**1 Introduction**

Soft robotics, driven by advancements in smart materials and actuation mechanisms, has gained attention due to its potential for safe, adaptable, and efficient interaction with unstructured environments [1, 2]. Among various actuation technologies, dielectric elastomer actuators (DEAs) emerged as a promising solution, offering high energy density, large deformation capability, and a lightweight structure [3]. DEAs can be configured into various forms, each enabling a distinct mode of motion, including swimming [4], rolling [5], crawling [6], hopping [7], and even flying [8], making them highly suitable for a diverse range of robotic locomotion applications. For example, circular DEAs with biasing mechanisms [9] or springs [10] can generate out-of-plane deformations, enabling both linear and rotational movement [11]. Saddle-shaped DEAs are widely employed in crawling soft robots [12], while cylindrical DEAs, capable of bending and linear elongation [13], are extensively used in artificial muscle applications.

Planar DEAs utilise a pre-stretched dielectric elastomer supported by a planar frame, enabling in-plane motion [14] such as linear displacement or rotation. Localised actuation in specific regions facilitates the development of multi-directional crawling robots [15]. When operated near their natural frequency, planar DEAs exhibit rapid and agile motion [16], making them particularly suitable for lightweight and flexible soft robotic systems. Compared with other DEA configurations, planar DEAs [17] offer the advantage of achieving in-plane deformations within a thin-profile structure, allowing them to access confined and complex environments [18]. However, despite these benefits, most DEA-driven in-plane soft robots still rely on tethered power sources due to the high voltage requirements of DEAs, which often exceed several kilovolts. This reliance on external wiring significantly limits mobility and operational flexibility, posing challenges for practical deployment in untethered applications.

In current DEA research, there are two primary approaches to achieving untethered actuation. The first involves miniaturising the actuation system, integrating a compact battery and high-voltage converter directly into the actuator [12]. While this method enables high-voltage actuation—often reaching kilovolt levels—it presents challenges in achieving lightweight and compact designs, as commercial high-voltage converters typically weigh more than ten grams, making integration difficult for small-scale soft robots [19]. The second approach focuses on reducing the required actuation voltage to the hundred-volt range by minimising the thickness of individual elastomer layers [20]. By employing a multi-layer stacking strategy [21], DEAs can still achieve high power output despite operating at a lower voltage. Simultaneously, this approach necessitates the development of lightweight and compact onboard electronics capable of delivering the required actuation voltage [22]. However, despite these advancements, the performance of untethered DEA-based soft robots remains limited. The motion capabilities of existing untethered DEA-driven systems are often simplistic, typically restricted to a single locomotion mode, highlighting the need for further improvements in multi-modal motion control and power efficiency.

Aiming to develop a compact, thin-profile untethered DEA-based soft locomotion robot, this study integrates a lightweight onboard electronics system (0.34 grams) with an in-plane thin dielectric elastomer actuator (TS-DEA) to enable wireless locomotion using three directional friction pads (Fig. 1). The TS-DEA features a dual-actuation sandwiched structure, consisting of four dielectric elastomer layers bonded to a compressible tensioning mechanism at its core. The robot measures 38 mm in length, 6 mm in height, and weighs just 2.34 grams, operating at a resonant frequency of 86 Hz under a low driving voltage of 220 V. Building upon existing research in in-plane DEA locomotion, this design demonstrates significant potential for wireless multimodal motion, enhancing the functional capabilities of soft robots for applications in constrained and dynamic environments.

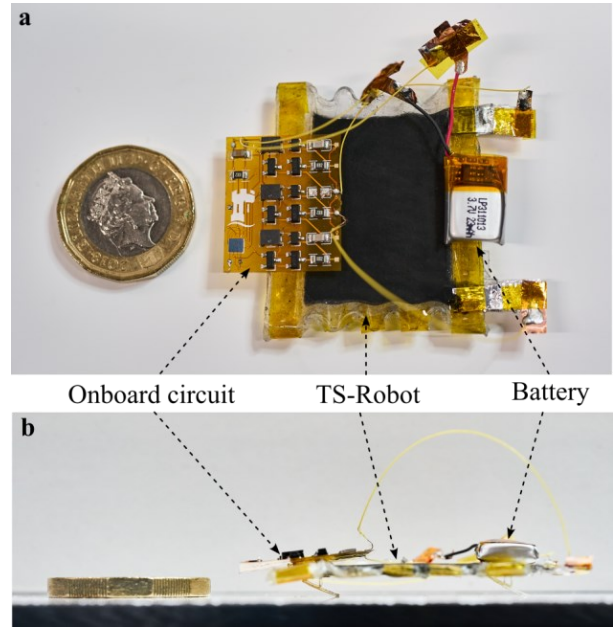

**Fig. 1 Prototype of the UTS-Robot.** (a) Top view; (b) Front view.

## 2 Untethered TS-Robot

The untethered TS-Robot (UTS-Robot) consists of onboard electronics (circuit and battery) and TS-Robot, as illustrated in Fig. 1. We used the same procedures presented in our previous work [18] to design the TS-Robots, which comprise two key subsystems: a thin, soft dielectric elastomer actuator (TS-DEA, Fig. 2a) for generating high-frequency displacement and driving forces, and three directional friction pads (Fig. 2b) for anchoring the robot to substrate surfaces. The TS-DEA adopts a dual-actuation sandwiched structure, consisting of four dielectric elastomer layers bonded to a compressible tensioning mechanism at the center. Each dielectric elastomer layer is fabricated using a spin coater (WS-650Mz-23NPPB, Laurell) with Ecoflex 0030 as the dielectric material. Initially, the elastomer is manufactured to a thickness of 50 μm and then biaxially pre-stretched by a factor of 2.5 × 2.5, resulting in a final thickness of 8 μm per layer. The elastomer is subsequently coated with multi-walled carbon nanotube electrodes (Thermo Scientific) before being bonded to the tensioning mechanism (1 mm PETG, laser-cut, Jindiao Technology Co., Ltd, JD3050) using an adhesive layer (Bostik Bond-Flex 100). This compressible tensioning mechanism enables in-plane deformation while maintaining a thin profile, optimising the actuator for compact applications.

The directional friction pads, which anchor the robot to the substrate, are laser-cut from 0.1 mm thick PET and folded into a triangular shape, as shown in Fig. 2b. The pad's inclined angle in contact with the ground is 30 degrees, providing effective frictional interaction for locomotion.

**Fig. 2 Design of the thin soft robot (TS-Robot).** (a) Thin soft dielectric elastomer actuator (TS-DEA); (b) Directional friction pad.

The onboard electronics comprise a flexible printed circuit board (FPCB) and a miniature battery. The FPCB, measuring 20 × 18 mm, integrates a microcontroller and a flyback converter, both powered by a 3.7V onboard battery (Fig. 3 a and b). The selected components are given in Table 1. As illustrated in Fig. 3c, the flyback circuit operates as follows: First, switch Q1 is activated via pulse-width modulation (PWM), allowing current to flow through the primary side of the transformer. When Q1 is switched off, the transformer releases its stored energy to the secondary side, generating a voltage and current pulse. The diode (D1) ensures unidirectional current flow, enabling capacitor C1 to accumulate a high voltage. Meanwhile, switch Q2 is responsible for discharging the DEA. The DEA can be actuated at a precisely controlled frequency by alternating charging and discharging cycles.

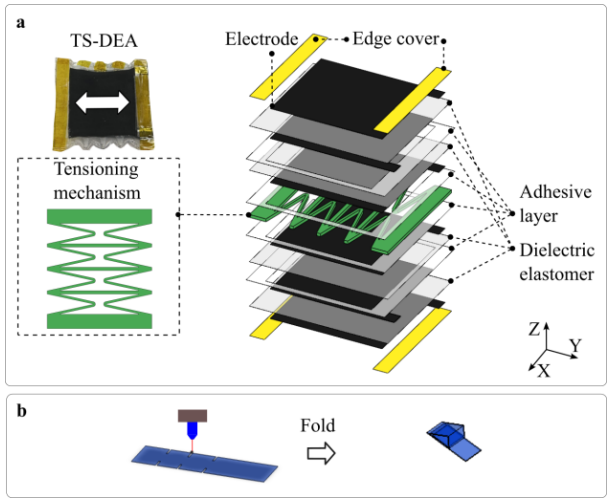

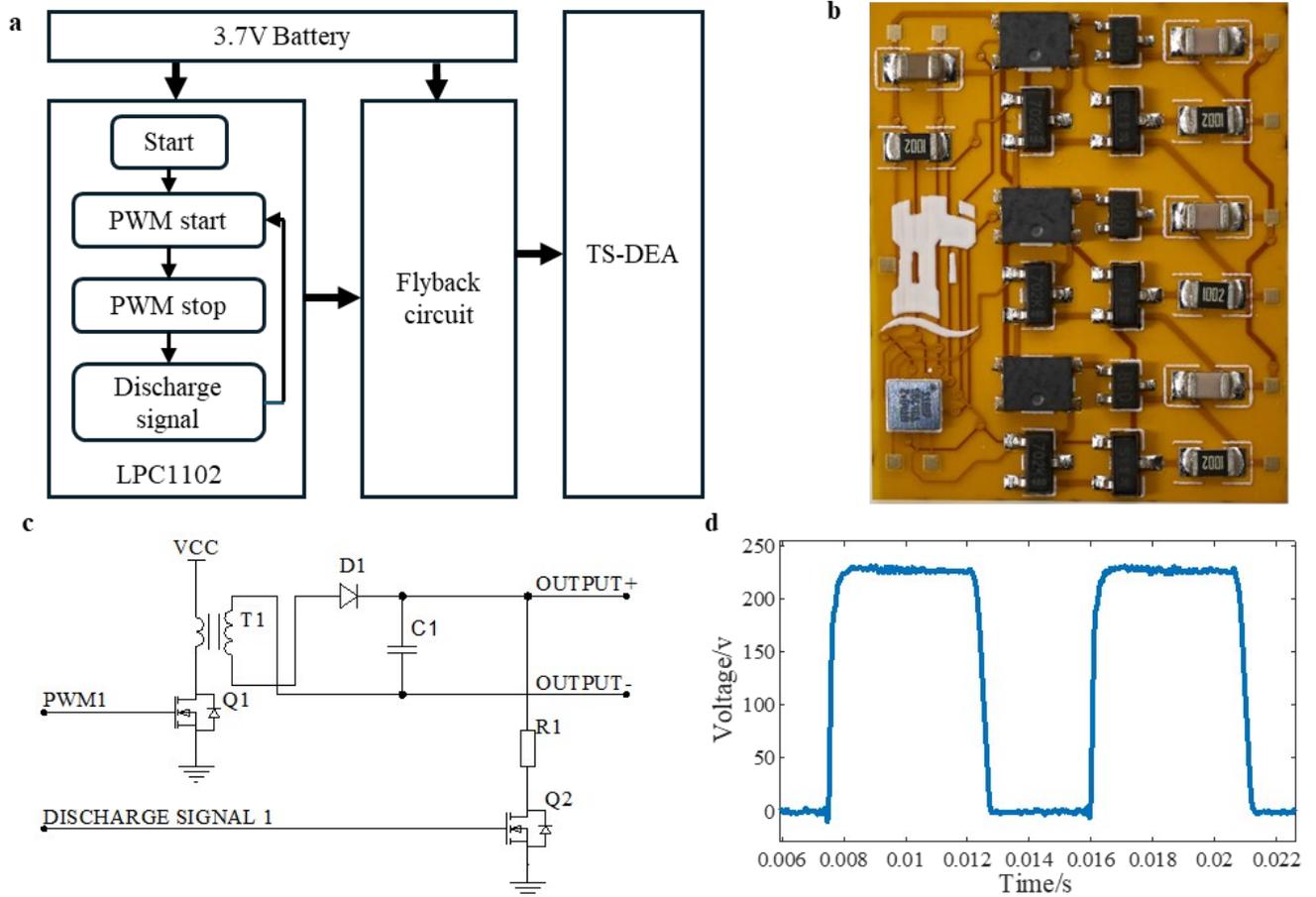

**Fig. 3 Onboard electronics for the TS-Robot.** (a) Block diagram of the onboard electronics workflow; (b) Prototype of the FPCB; (c) Single channel flyback circuit for driving TS-DEA; (d) Example of the circuit output test.

The electronics feature output channels capable of generating waveforms with amplitudes of up to 220 V and frequencies of up to 1 kHz as exampled in Fig. 3d. These parameters can be dynamically adjusted by the onboard microcontroller, enabling precise control and efficient operation of the dielectric elastomer actuators. The current onboard electronics (Fig. 3b) are designed with three independent output channels, allowing for untethered, multimodal locomotion in DEA-driven thin soft robots. However, for this study, only a single channel was utilised to drive a single TS-Robot for demonstration.

**Table 1 Components of the onboard electronics shown in Fig. 3b**

| Description | Model | Quantity |
| --- | --- | --- |
| Capacitor | 10uF | 1 |
| Capacitor | 100pF | 3 |
| Resistor | 10kΩ | 4 |
| Diode | CMSD2004s | 3 |
| MOSFET | 2N7002K | 3 |
| MOSFET | BSS127S-7 | 3 |
| Transformer | ATB322515-0110 | 3 |
| Controller | LPC1102UK,118 | 1 |
| FPCB | - | 1 |
| Battery | LP311013 | 1 |

## 3 Modelling

We proposed the model for TS-DEA to calculate the voltage-induced displacement and natural frequency. As illustrated in Fig. 2a, the TS-DEA is constructed by four dielectric elastomer layers and an in-plane compressible tensioning mechanism. The parameters used for the modal is presented in Table 2. The original dimension of each dielectric elastomer layer is $L_{a\_x}$, $L_{a\_y}$, and $L_{a\_z}$ in the X, Y and Z axes. It is then pre-stretched biaxially to $l_{a\_xpre}$, $l_{a\_ypre}$, and $l_{a\_zpre}$ respectively. Then adhered to the tensioning mechanism (original length $L_{t\_y}$) through glue. The force balance between the tensioning mechanism and actuation layer results in the free state TS-DEA where the tensioning mechanism is compressed to $l_{t\_y}$ and the length of the un-actuated TS-DEA is represented by $l_{DEA} = l_{t\_y} = l_{a\_y}$. The deformation ratio of the dielectric elastomer layer is represented by $\lambda_{a\_x} = l_{a\_xpre}/L_{a\_x}$, $\lambda_{a\_y} = l_{a\_y}/L_{a\_y}$ and respectively, where $\lambda_{a\_x}\lambda_{a\_y}\lambda_{a\_z} = 1$.

Based on the Gent model [23], the stress and the force of the actuation layer in Y axis is:

$$\sigma_{a\_y} = \mu_1 \frac{\lambda_{a\_y}^2 - \lambda_{a\_x}^{-2}\lambda_{a\_y}^{-2}}{1 - \frac{\lambda_{a\_x}^2 + \lambda_{a\_y}^2 + \lambda_{a\_x}^{-2}\lambda_{a\_y}^{-2} - 3}{J_1}} \quad (1)$$

$$F_{a\_y} = 4\sigma_{a\_y}A_{a\_y} \quad (2)$$

where $A_{a\_y} = l_{a\_x}l_{a\_z}$ is area of the single actuation layer in Y axis. Therefore, the stiffness of the actuation layer is represented by

$$k_{a\_y} = \frac{4A_{a\_y}\sigma_{a\_y}}{l_{a\_y} - L_{a\_y}} \quad (3)$$

The voltage induced force in Y axis is

$$F_{v\_y} = \frac{4\varepsilon U^2 \lambda_{a\_x}^2 l_{a\_x}l_{a\_z}}{L_{a\_z}^2} \quad (4)$$

where the $\varepsilon$ is the dielectric constant of the elastomer, $U$ is the applied voltage. The deformation of the tensioning mechanism is modelled by compliant mechanism [18] and the reaction force is expressed as

$$k_{t\_y} = \frac{3EI_t\cos\theta}{2A_{t\_y}l^2} \quad (5)$$

$$F_{t\_y} = k_{t\_y}(L_{t\_y} - l_{a\_y}) \quad (6)$$

where $A_{t\_y} = t_t h$ is area of the tensioning mechanism in the Y axis, $h$ is the width in the X axis, $t_t$ is the thickness in the Z axis. $E$ is the elastic modulus of the tensioning mechanism material, $I_t$ is the second moment of area of the inclined linkage cross section, $I_t = t_t w^3/12$, $w$ is the width of the inclined linkage. $\theta$ is the angle between the inclined linkage and the Y axis, and $l$ is the length of the inclined linkage.

The length of the TS-DEA in un-actuated state $l_{DEA}$ can be obtained by $F_{t\_y} = F_{a\_y}$. The voltage-induced displacement is calculated through $F_{t\_y} + F_{v\_y} = F_{a\_y}$. The high-frequency response of the in-plane DEA can be simplified as a mass-spring-damper system, and the natural frequency of the in-plane DEA can be approximated by ($m$ is the mass of the TS-DEA):

$$f = \frac{1}{2\pi}\sqrt{\frac{k_{a\_y} + k_{t\_y}}{m}} \quad (7)$$

**Table 2 Parameters of the TS-DEA**

| | |
| --- | --- |
| $L_{a\_x} * L_{a\_y} * L_{a\_z}$ | 16*14*0.05mm |
| $\lambda_{a\_x} * \lambda_{a\_y}$ | 2.5*2.5 |
| $L_{t\_y}$ | 35mm |
| $\mu_1$ | 23000 |
| $J_1$ | 97 |
| $E$ | 2.4Gpa |
| $t_t$ | 1mm |
| $w$ | 0.3mm |
| $\theta$ | 77.8° |
| $m$ | 1.2g |
| $l$ | 16.3mm |
| $h$ | 40mm |
| $\varepsilon$ | 3* 8.85 × 10$^{-12}$ |

# 4 Characterisation

## 4.1 Displacement and blocking force of the TS-DEA

In this section, the static displacement, blocking force, and displacement amplitude of the TS-DEA at different actuation frequencies are characterised. The actuation signals are generated using a signal generator (WaveStation 2022, Teledyne LeCroy) and amplified by a factor of 1000 using high-voltage amplifiers (Model 610E-K-CE, Trek Inc.). A stair wave signal characterises the static displacement and blocking force, while a sweep frequency signal evaluates displacement performance across various frequencies. A laser sensor (LK-031, Keyence) and a load cell (LCFD-1KG, Omega) record displacement and force data, respectively.

Fig. 4 a and b present the static displacement and blocking force, respectively, both of which increase with applied voltage. The maximum voltage applied during the measurements is 220 V, corresponding to an electric field strength of 27.5 V/μm. The maximum static displacement and blocking force recorded are 82 μm and 15 mN, respectively. As illustrated in Fig. 4c, when actuated at 220 V with frequencies ranging from 1 Hz to 150 Hz, the displacement response of the TS-DEA initially decreases with increasing frequency, reaching a peak at 86 Hz, which is estimated as the first-order resonant frequency of the TS-DEA.

## 4.2 Electrical calculation of the TS-DEA for the high-frequency actuation

After fabricating the TS-DEA, the resistance and capacitance were measured using the ModuLab XM system, with values averaged over frequencies ranging from 50 to 100 Hz, resulting in 0.41 MΩ and 1.69 nF, respectively. To achieve high-frequency actuation, the electrical cutoff of the TS-DEA can be estimated using the RC time constant as:

$$f_c = \frac{1}{2\pi RC} \quad (8)$$

where $f_c$ is the electrical cutoff frequency, $R$ is the total electrical resistance of the electrodes, and $C$ is the capacitance of the TS-DEA. This calculation results in an electrical cutoff frequency of 230 Hz, which sufficiently covers the first-order resonant frequency, as illustrated in Fig. 4c.

The energy efficiency of the TS-DEA was evaluated based on the methodology proposed in [24, 25]. The efficiency of a general actuator is defined as:

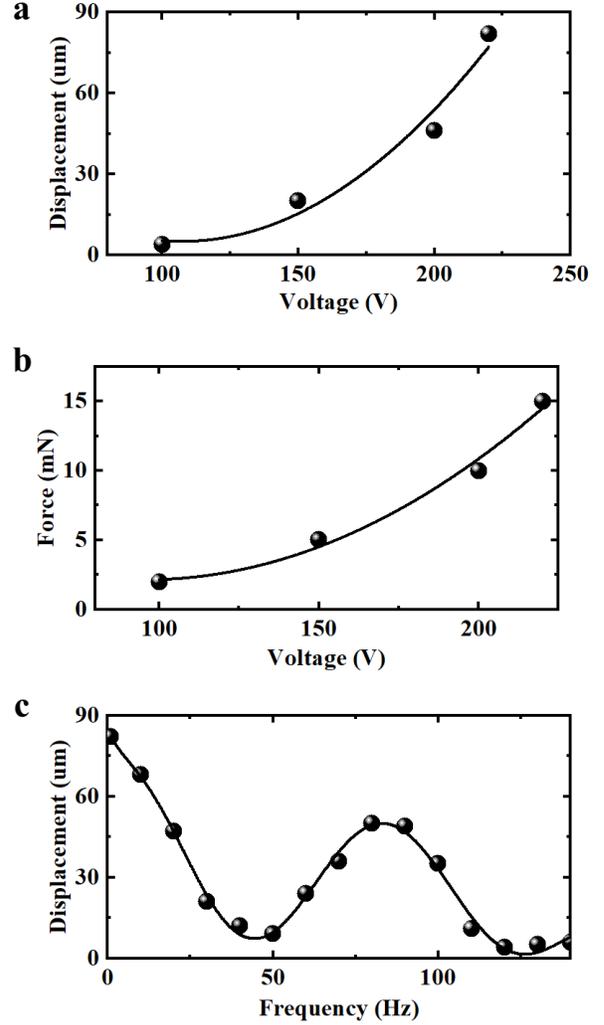

**Fig. 4 Characterisation result of the TS-DEA.** (a) Displacement vs Voltage; (b) Blocking force vs Voltage; (c) Displacement amplitude vs Frequency.

$$\eta = \frac{P_{DEA}}{P_{in}} \quad (9)$$

where $P_{DEA}$ is the mechanical power generated by the TS-DEA, and $P_{in}$ is the input electrical power. Assuming a linear approximation, $P_{DEA}$ can be expressed as:

$$P_{DEA} = \frac{1}{2} F_B \Delta f \quad (10)$$

where $F_B$ and $\Delta$ represent the blocking force and voltage-induced displacement of the TS-DEA at an actuation frequency $f$. The input power $P_{in}$ is determined by integrating the time-dependent voltage $U$ and current $I$ at frequency $f$:

$$P_{in} = f \int_0^T UI dt \quad (11)$$

where $T$ is the period of a single actuation cycle. We utilised the measurement device from our previous work [18] to determine the input power ($U$ and $I$) of the TS-DEA. Focusing on the resonant frequency, the mechanical power output of the TS-DEA was measured as 0.03 mW, resulting in energy efficiency of 0.1%.

## 5 Demonstration of the Bare TS-Robot and untethered TS-Robot

To effectively evaluate the locomotion performance of the untethered TS-Robot, we first assessed the locomotion capability of the bare TS-Robot when actuated by an external power system, consisting of a signal generator (WaveStation 2022, Teledyne LeCroy) and high-voltage amplifiers (Model 610E-K-CE, Trek Inc.). Subsequently, we demonstrated the locomotion of the UTS-Robot driven by the onboard electronics.

The bare TS-Robot was actuated using a 220 V, 86 Hz square wave, as illustrated in Fig. 5a and supplementary video. The locomotion tests were conducted on an acrylic sheet substrate, where the robot achieved a maximum crawling speed of 12.36 mm/s, equivalent to 0.33 body lengths per second and 3.75 body thicknesses per second. Additionally, we evaluated the load-carrying capability of the bare TS-Robot, with results presented in Table 3. The locomotion speed decreased as the payload increased. When carrying a 3 g payload, the robot's speed dropped to 1.1 mm/s, corresponding to 0.03 body lengths per second and 0.33 body thicknesses per second. These results indicate that locomotion speed is highly sensitive to the payload, primarily due to changes in the contact conditions between the directional friction pad and the substrate, which subsequently affect the effective movement distance per locomotion cycle.

We then tested the locomotion capability of the untethered TS-Robot (UTS-Robot), as illustrated in Fig. 5b and supplementary video. The actuation signal was set to 220 V, 86 Hz square wave, identical to that used for the bare TS-Robot. The results demonstrated that the UTS-Robot moved on a polyimide substrate at a speed of 0.5 mm/s, equivalent to 0.015 body lengths per second and 0.083 body thicknesses per second. From the experiment, we noticed that TS-Robots with directional friction pads exhibit faster locomotion on smoother surfaces. Additionally, the untethered TS-Robot actuated by onboard electronics shown a lower locomotion speed compared to the bare TS-Robot actuated by the external Trek 610E power source. The main factor is that the TS-DEA consists of four elastomer layers, each with an area of 40 mm × 35 mm, resulting in high capacitance. Therefore, the onboard electronics unable to fully charge and discharge the DEA within each actuation cycle, limiting its performance.

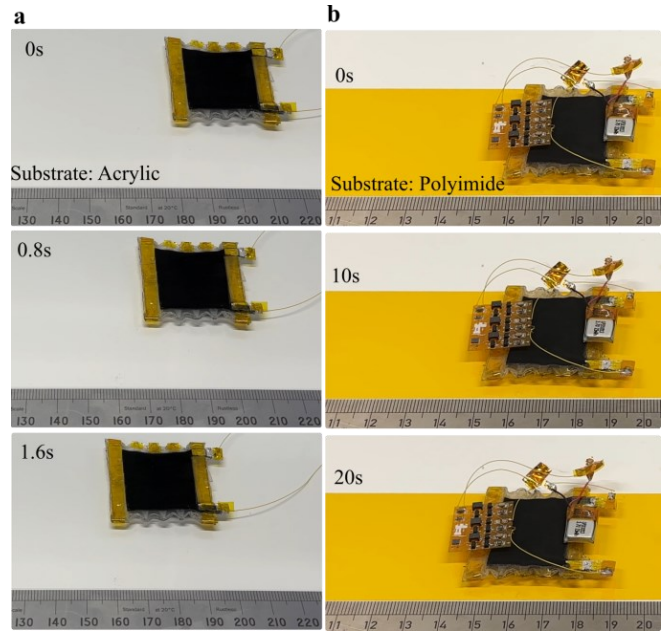

**Fig. 5 Demonstration of the bare and untethered TS-Robot (Supplimentary video).** (a) Bare TS-Robot; (b) Untethered TS-Robot.

**Table 3 Bare TS-Robot speed versus payload**

| Payload (g) | Speed (mm/s) | Speed (Body Length/s) | Speed (Body Thickness/s) |
|---|---|---|---|
| 0 | 12.36 | 0.33 | 3.75 |
| 0.25 | 4.79 | 0.13 | 1.45 |
| 0.5 | 4.68 | 0.13 | 1.42 |
| 1 | 3.97 | 0.11 | 1.20 |
| 1.5 | 2.21 | 0.06 | 0.67 |
| 1.75 | 2.06 | 0.06 | 0.62 |
| 2 | 1.9 | 0.05 | 0.58 |
| 2.5 | 1.8 | 0.05 | 0.55 |
| 3 | 1.1 | 0.03 | 0.33 |

## 6 Discussion and Conclusion

This study presents a novel untethered thin soft robot (UTS-Robot), addressing key challenges in soft robotics, achieving low-voltage, low profile and wireless locomotion. A novel in-plane soft dielectric elastomer actuator (TS-DEA) was designed, enabling compact, high-frequency actuation while maintaining a low-profile structure. The dual-actuation sandwiched design ensures in-plane deformation, which is crucial for locomotion in confined environments. By employing spin coating techniques, we achieved a reduced dielectric elastomer thickness, allowing for low-voltage actuation at 220 V, making them more practical for untethered applications.

The onboard electronics were developed with a flexible printed circuit board (FPCB) integrating a flyback converter and a microcontroller. The system enables wireless actuation with output waveforms up to 220 V and 1 kHz, a critical step toward autonomous soft robotics. A comprehensive model was established to predict voltage-induced displacement and natural frequency, providing a theoretical framework for optimising DEA performance. The bare TS-Robot, actuated by an external power supply, achieved a crawling speed of 12.36 mm/s, demonstrating the effectiveness of the TS-DEA design. The untethered TS-Robot (UTS-Robot), driven by the onboard electronics, successfully achieved wireless locomotion on a polyimide substrate at a speed of 0.5 mm/s.

The UTS-Robot moves at a slower speed compared to the externally actuated bare TS-Robot. This is due to the large capacitance of the TS-DEA, which consists of four elastomer layers (40 mm × 35 mm each). The increased capacitance limits the onboard electronics' ability to fully charge and discharge the actuators during high-frequency actuation, reducing overall efficiency. Additionally, the onboard power supply is less effective than the high-voltage amplifier (Trek 610E) used for the bare TS-Robot, resulting in lower actuation energy and slower response speed. To enhance untethered locomotion performance, future optimisation of the TS-DEA design will focus on reducing capacitance, for example, by decreasing the actuator area to improve charging efficiency.

While the current design enables basic untethered movement, future work will focus on developing multimodal locomotion (e.g., steering, crawling, and transitioning) and advancing multi-functional untethered soft robots capable of operating in complex and constrained environments.

## Acknowledgment

This work was supported by the China Scholarship Council Research Excellence Scholarship with the University of Nottingham (Dr Xi Wang, Mr Siqian Li), EPSRC (EP/W001128/1) and Innovate UK (51689).

## Declarations

**Competing Interests** The authors declare that they have no competing interests.

## References

1. H. Hussein *et al.* (2023 Jul 2023). Actuation of Mobile Microbots: A Review. *Adv. Intell. Syst.,* p. 25. 10.1002/aisy.202300168
2. S. Zhang, X. X. Ke, Q. Jiang, Z. P. Chai, Z. G. Wu, and H. Ding. (Dec 2022). Fabrication and Functionality Integration Technologies for Small-Scale Soft Robots. *Advanced Materials,* vol. 34, no. 52, p. 32. 10.1002/adma.202200671
3. Y. G. Guo, L. W. Liu, Y. J. Liu, and J. S. Leng. (Oct 2021). Review of Dielectric Elastomer Actuators and Their Applications in Soft Robots. *Adv. Intell. Syst.,* vol. 3, no. 10, p. 18. 10.1002/aisy.202000282
4. J. Shintake, V. Cacucciolo, H. Shea, and D. Floreano. (Aug 2018). Soft Biomimetic Fish Robot Made of Dielectric Elastomer Actuators. *Soft Robot,* vol. 5, no. 4, pp. 466-474. 10.1089/soro.2017.0062
5. O. A. Araromi *et al.* 2015). Rollable Multisegment Dielectric Elastomer Minimum Energy Structures for a Deployable Microsatellite Gripper. *IEEE/ASME Transactions on Mechatronics,* vol. 20, no. 1, pp. 438-446. 10.1109/tmech.2014.2329367
6. G. Gu, J. Zou, R. Zhao, X. Zhao, and X. Zhu. 2018). Soft wall-climbing robots. *Science Robotics,* vol. 3, no. 25.
7. M. Duduta, F. Berlinger, R. Nagpal, D. R. Clarke, R. J. Wood, and F. Z. Temel. (Jul 2020). Tunable Multi-Modal Locomotion in Soft Dielectric Elastomer Robots. *Ieee Robotics and Automation Letters,* vol. 5, no. 3, pp. 3868-3875. 10.1109/lra.2020.2983705
8. S. Kim, Y.-H. Hsiao, Z. Ren, J. Huang, and Y. Chen. 2025). Acrobatics at the insect scale: A durable, precise, and agile micro–aerial robot. *Sci. Robot.,* vol. 10, no. 98, p. eadp4256.
9. P. Loew, G. Rizzello, and S. Seelecke. (Dec 2018). A novel biasing mechanism for circular out-of-plane dielectric actuators based on permanent magnets. *Mechatronics,* vol. 56, pp. 48-57. 10.1016/j.mechatronics.2018.10.005
10. C. J. Cao, T. L. Hill, B. Li, L. Wang, and X. Gao. (Jun 2021). Nonlinear dynamics of a conical dielectric elastomer oscillator with switchable mono to bi-stability. *International Journal of Solids and Structures,* vol. 221, pp. 18-30. 10.1016/j.ijsolstr.2020.02.012
11. R. Wache, D. N. McCarthy, S. Risse, and G. Kofod. (Apr 2015). Rotary Motion Achieved by New Torsional Dielectric Elastomer Actuators Design. *IEEE-ASME Trans. Mechatron.,* vol. 20, no. 2, pp. 975-977. 10.1109/tmech.2014.2301633
12. J. W. Cao *et al.* (May 2018). Untethered soft robot capable of stable locomotion using soft electrostatic actuators. *Extreme Mechanics Letters,* vol. 21, pp. 9-16. 10.1016/j.eml.2018.02.004
13. C. Tang *et al.* 2022). A pipeline inspection robot for navigating tubular environments in the sub-centimeter scale. *Science Robotics,* vol. 7, no. 66, p. eabm8597. doi:10.1126/scirobotics.abm8597
14. J. L. Guo, C. Q. Xiang, A. Conn, and J. Rossiter. (Jun 2020). All-Soft Skin-Like Structures for Robotic Locomotion and Transportation. *Soft Robotics,* vol. 7, no. 3, pp. 309-320. 10.1089/soro.2019.0059
15. K. M. Digumarti, C. J. Cao, J. L. Guo, A. T. Conn, J. Rossiter, and Ieee. 2018). Multi-directional Crawling Robot with Soft Actuators and Electroadhesive Grippers. *2018 Ieee International Conference on Soft Robotics (Robosoft),* pp. 303-308. [Online]. Available: <Go to ISI>://WOS:000610424800049
16. E. M. Henke, S. Schlatter, and I. A. Anderson. (Dec 2017). Soft Dielectric Elastomer Oscillators Driving Bioinspired Robots. *Soft Robot,* vol. 4, no. 4, pp. 353-366. 10.1089/soro.2017.0022
17. X. Wang, L. Raimondi, D. Axinte, and X. Dong. 2024). Investigation on a class of 2D profile amplified stroke dielectric elastomer actuators. *Journal of Mechanisms and Robotics,* pp. 1-32.
18. X. Wang, S. Li, J.-C. Chang, J. Liu, D. Axinte, and X. Dong. 2024). Multimodal locomotion ultra-thin soft robots for exploration of narrow spaces. *Nature Communications,* vol. 15, no. 1, p. 6296.
19. N. Sriratanasak, D. Axinte, X. Dong, A. Mohammad, M. Russo, and L. Raimondi. (Dec 2022). Tasering Twin Soft Robot: A Multimodal Soft Robot Capable of Passive Flight and Wall Climbing. *Adv. Intell. Syst.,* vol. 4, no. 12, p. 13. 10.1002/aisy.202200223
20. X. Ji *et al.* 2019). An autonomous untethered fast soft robotic insect driven by low-voltage dielectric elastomer actuators. *Sci. Robot.,* vol. 4, no. 37, p. eaaz6451. doi:10.1126/scirobotics.aaz6451
21. H. C. Zhao, A. M. Hussain, M. Duduta, D. M. Vogt, R. J. Wood, and D. R. Clarke. (Oct 2018). Compact Dielectric Elastomer

| | |
|---|---|
| | Linear Actuators. *Adv. Funct. Mater.,* vol. 28, no. 42, p. 12. 10.1002/adfm.201804328 |
| 22 | F. Hartmann, M. Baskaran, G. Raynaud, M. Benbedda, K. Mulleners, and H. Shea. 2025). Highly agile flat swimming robot. *Sci. Robot.,* vol. 10, no. 99, p. eadr0721. |
| 23 | G.-Y. Gu, U. Gupta, J. Zhu, L.-M. Zhu, and X. Zhu. 2017). Modeling of Viscoelastic Electromechanical Behavior in a Soft Dielectric Elastomer Actuator. *IEEE Transactions on Robotics,* vol. 33, no. 5, pp. 1263-1271. 10.1109/tro.2017.2706285 |
| 24 | J. P. L. Bigué and J. S. Plante. (Feb 2013). Experimental Study of Dielectric Elastomer Actuator Energy Conversion Efficiency. *Ieee-Asme Transactions on Mechatronics,* vol. 18, no. 1, pp. 169-177. 10.1109/tmech.2011.2164930 |
| 25 | Y. Chen *et al.* (Nov 2019). Controlled flight of a microrobot powered by soft artificial muscles. *Nature,* vol. 575, no. 7782, pp. 324-329. 10.1038/s41586-019-1737-7 |